\documentclass{edm_article}
\usepackage{natbib}
\usepackage{algorithmicx}
\usepackage{algpseudocode}
\usepackage[linesnumbered,ruled,vlined]{algorithm2e}
\usepackage{float}
\usepackage{pgfplots}
\pgfplotsset{compat=1.18}
\usepgfplotslibrary{groupplots}
\usepackage{tabularx}
\usepackage{makecell}
\usepackage{booktabs}     
\usepackage{subcaption}   
\usepackage{multirow}
\usepackage{amsmath}
\usepackage[normalem]{ulem}

\newcommand{\revision}[1]{\textcolor{black}{#1}}

\newcommand{\revisiontwo}[1]{\textcolor{black}{#1}}

\newcommand{\revisionrd}[1]{\textcolor{black}{#1}}

\newcommand{\revisioncamera}[1]{\textcolor{black}{#1}}

\begin{document}

\title{From Heuristics to Analytics: Forecasting Effort and Progress in Online Learning}


\numberofauthors{5}
\author{
\alignauthor
Eric S. Qiu\\
       \affaddr{Cornell University}\\
       \affaddr{Ithaca, NY, USA}\\
       \email{sq225@cornell.edu}
\alignauthor
Danielle R. Thomas\\
       \affaddr{Carnegie Mellon University}\\
       \affaddr{Pittsburgh, PA, USA}\\
       \email{drthomas@cmu.edu}
\alignauthor
Boyuan Guo\\
       \affaddr{Carnegie Mellon University}\\
       \affaddr{Pittsburgh, PA, USA}\\
       \email{boyuang@andrew.cmu.edu}
\and  
\alignauthor
Vincent Aleven\\
       \affaddr{Carnegie Mellon University}\\
       \affaddr{Pittsburgh, PA, USA}\\
       \email{aleven@cs.cmu.edu}
\alignauthor
Conrad Borchers\\
       \affaddr{Carnegie Mellon University}\\
       \affaddr{Pittsburgh, PA, USA}\\
       \email{cborcher@cs.cmu.edu}
}

\maketitle

\begin{abstract}
\revision{
Sustained effort is essential for realizing the benefits of intelligent tutoring systems (ITS), yet many learners disengage or underuse available practice time. We introduce \emph{engagement forecasting} as a supervised prediction task based on ITS logs, targeting two outcomes central to effort and learning \revisiontwo{progress}: \textbf{minutes practiced per week} and \textbf{new skills mastered per week}. Using interaction log data from 425 middle-school students over a school year, we benchmark fifteen predictors including regressions, decision trees, and neural networks\revisiontwo{. We} show that these feature-based models reduce mean absolute error \revision{(MAE)} by 22--33\% relative to heuristic baselines, including fixed-percentile rules adapted from prior work in other behavioral domains. We find that percentile heuristics systematically overpredict, whereas feature-based models better track student practice trajectories across weeks. To support explainability, we analyze feature importance and ablations, revealing target-specific patterns: effort forecasting is driven mainly by recent activity features, while progress forecasting depends more on learner-state and content difficulty signals. 
\revisiontwo{Finally, in a \revisionrd{semi-structured user interview case study} with eight college tutors, we examine how tutors reasoned about system-generated predictive features when setting goals with students. We find that tutors reasoned differently about effort versus progress goals in ways that mirror our pattern analysis.
\revisionrd{Together, these results establish a reproducible benchmark for forecasting weekly effort and learning progress in ITS. By making patterns of sustained effort and progress visible at a weekly timescale, engagement forecasting offers a foundation for supporting tutor–learner goal setting and timely instructional decisions.}
} 
}
\end{abstract}


\section{\revision{Introduction}}
Active learning systems that scaffold student problem solving can substantially improve learning outcomes, especially when learners engage in high volumes of practice with feedback over several months \cite{ma2014intelligent,kulik2016effectiveness}. Yet, persistence and effort remain bottlenecks: many students disengage after initial practice or do not utilize the class time they have at their disposal, limiting these systems' long-term impact \cite{gurung2025starting,koedinger2015learning}. In large-scale deployments, engagement is often highly skewed---for example, only a small minority of students may reach recommended weekly usage levels \cite{holt2024fivepercent, grimaldi2022estimating, eames2026khan}. This disparity in engagement motivates a central question: how can analytics predict and explain differences in student effort to generate actionable, learner- and teacher-facing insights that support effort regulation?

In this paper, we introduce \emph{engagement forecasting}: a supervised prediction task that estimates a learner's next-week engagement level from prior ITS interaction traces. We focus on two engagement indicators that are both observable in ITS logs and directly relevant to effort regulation support: \textbf{minutes practiced per week} (effort) and \textbf{new skills mastered per week} (learning progress) \cite{schaldenbrand2021computer}. 
\revisionrd{We adopt these targets not as complete measures of motivation or learning, but as simple, observable signals that tutors and platforms already use for week-to-week planning \cite{grimaldi2022estimating}, making them suitable for decision support even when imperfect.}

\revisiontwo{Importantly, student engagement in ITS is constrained, but not fully fixed. Middle-school students typically engage with tutoring software within clear boundaries set by teachers and schools, such as designated class time and homework policies. Yet, students still exercise agency within these constraints: how consistently they use allotted time, whether they persist across weeks, and how they respond to prior success or difficulty. This motivates engagement forecasting at an intermediate (e.g. weekly) time scale: it can serve as actionable decision-support signals for teachers and students, helping guide agency and self-regulation as students progress through the school year \cite{SRL}.}

\revisiontwo{A key challenge in forecasting engagement is identifying which aspects of learner context matter. Motivational theory suggests that persistence is shaped by beliefs about competence (often discussed as self-efficacy) as well as recent success and failure experiences \cite{ryan2000,vicious}. From an analytics perspective, some of these factors are indirectly observable in ITS logs through patterns in learner interactions. For example, difficulty and learning rate—long studied in EDM through models such as the Additive Factors Model (AFM)—may influence whether continued effort translates into progress or discouragement \cite{cen2006}. Rather than posing a new theory of motivation, we seek empirical insights into which of these log-derived signals are most predictive of next-week engagement targets.}

We compare feature-driven predictive models against strong heuristic baselines. In addition to common algorithms (e.g., mean and median), we evaluate percentile rules adapted from adjacent health behavior domains \cite{2x2,dashboardgoalsetting}. These percentile heuristics often assume sustained week-over-week improvement; whether that assumption transfers to K--12 tutoring settings is an open question \cite{gurung2025starting}. \revision{We further conduct feature attribution and ablation tests across four feature groups derived from ITS logs: AFM-based learner state, recent activity trends, practice gaps, and prior achievement. By evaluating these groups, we seek to identify which contextual signals are most \revisiontwo{predictive} of student effort, potentially using these insights to improve future ITS adaptivity \cite{Xia2025OptimizingML}. 
\revisionrd{To move towards closing this loop, we conduct a semi-structured user interview to examine how tutors can use the explainable signals when setting goals with students.}
} We ask the following research questions:

\textbf{RQ1:} \revision{How well can data-driven approaches improve forecasting of next-week effort and progress}?
\newline
\textbf{RQ2:} Do percentile heuristics drawn from other contexts (e.g., health and sports) generalize to engagement forecasting in K--12 tutoring settings?
\newline
\textbf{RQ3:} Which features available from ITS data are most indicative of next-week effort and progress?

\revisionrd{
This paper contributes to EDM theoretically and practically.
First, we introduce week-ahead engagement forecasting in ITS as a new prediction task, focusing on calendar-level effort and progress to support practical instructional decision-making, complementing prior EDM work such as affect detection, wheel-spinning, and dropout prediction. With \textbf{RQ1} and \textbf{RQ2}, we show that statistical learning is a valid approach to the task that outperforms and better tracks student trajectory compared to existing heuristics.
Second, through feature analysis in \textbf{RQ3}, we provide theoretical insights into the structure of engagement, showing that effort and progress are predicted by qualitatively different features (recent activity vs. learner-centered signals). 
Third, we use a user interview case study to examine how these predictive signals align with tutors’ reasoning about engagement and goal setting, moving towards closing the loop between engagement forecasting and improving learning benefits.
}

\section{\revision{Related Work}}

We review prior work in (1) modeling effort, engagement, and disengagement in ITS; (2) \revisiontwo{foundational accounts of persistence and effort regulation; and (3) prediction tasks in EDM.} Across these areas, prior work motivates why effort is a distinct and consequential target, but also highlights that personalized, data-driven forecasting of week-ahead effort and progress remains underexplored.

\subsection{Effort and Engagement Modeling in ITS}

ITS generate fine-grained logs of student practice that enable automated feedback and data-driven analysis \cite{aleven2025integratedplatformstudyinglearning,vanlehn2011relative}. While ITS can improve learning outcomes, their impact is often constrained by the lack of sustained student engagement over long time horizons \cite{holt2024fivepercent,goalsetting1}. For example, large-scale deployments can exhibit substantial skew in usage, with many learners falling below recommended practice levels \revisiontwo{\cite{grimaldi2022estimating, eames2026khan}}. This motivates analytics that focus not only on \emph{knowledge progress} but also on \emph{effort and persistence} as first-class targets.

A substantial body of ITS research has modeled engagement-related phenomena such as off-task behavior, gaming the system, and other forms of disengagement using interaction traces (e.g., response patterns, hint usage, timing signals, delayed starting), often to trigger real-time supports \revision{including adaptive prompts or teacher alerts \cite{baker2004misuse, baker2007offtask, gurung2025starting}}. These works emphasize that observed (low) performance can reflect multiple latent causes (e.g., low effort, unproductive strategies) that can be detected by various engagement signals. Studies have also explored engagement-sensitive tutoring and motivational supports within tutors. \revision{For example, Arroyo et al.~\cite{arroyo2010effortbased} proposed and evaluated effort-based tutoring that infers effort from behavior and adapts activity selection accordingly. \revisiontwo{Similarly, the AnimalWatch ITS modeled student performance based on prior effort and observed learning gains from use}~\cite{beal2010animalwatch}.}

However, much of the literature operates at the level of problem-steps, sessions, or within-day interactions, and often evaluates detectors or interventions in-situ \cite{gurung2025starting}. In contrast, many educational decisions about pacing, check-ins, and expectations occur on weekly rhythms (e.g., weekly assignments, classroom schedules, weekly reflections). Our work complements fine-grained engagement modeling by introducing and benchmarking a \emph{week-ahead forecasting} task: predicting next-week outcomes from prior logs. This framing targets longer-term engagement rather than moment-to-moment disengagement detection.

\subsection{Motivation, Self-Efficacy, and Persistence}

Motivational theory provides a lens for why learners persist (or fail to persist) even when instructional support is available. Self-determination theory emphasizes the roles of competence, autonomy, and relatedness in sustaining motivation \cite{ryan2000}. Goal-setting theory further argues that specific and challenging goals can mobilize effort, but only when goals are perceived as attainable; repeated failure can undermine perceived competence and persistence \cite{lockelatham2002,LockeLatham2019}. This also aligns with accounts of vicious cycles in learning, where recent success and failure experiences can compound over time, shaping subsequent engagement and outcomes \cite{vicious}.

Empirically, motivational constructs such as self-efficacy have also been studied in learning contexts, including variability over time and relations to behavior and performance. \revision{For example, Bernacki et al.\ assessed self-efficacy and found that efficacy varied reliably during learning and predicted subsequent help-seeking and performance \cite{bernacki2015selfefficacy}.} These perspectives motivate our week-ahead targets in two ways. First, they suggest that effort and persistence are not static traits but can fluctuate with recent experiences and context, making forecasting meaningful at intermediate time scales (e.g., week-ahead). Second, they motivate evaluating forecasting not only as an error-minimization exercise, but as a foundation for decision-support while preserving learner agency. Importantly, our contribution is not to measure motivational states directly. Instead, we forecast behavioral and progress outcomes that are plausibly influenced by such dynamics, and we interpret feature patterns in light of these theories.

\subsection{\revisionrd{Student Behavior Prediction in EDM}}
\revisiontwo{Prior work in EDM has extensively modeled various prediction targets, but along dimensions that differ from our engagement forecasting task. For example, much of the ITS-focused literature operates at fine temporal granularity (problem-step or within-session windows), predicting affective states such as concentration, boredom, or frustration \cite{baker2012sensorfree, zambrano2024affect}, and detecting behaviors such as gaming the system or off-task activity. These studies typically define engagement as an affective construct (Engaged Concentration) inferred from ITS interaction traces, and the tasks are framed as supervised binary classification aimed at real-time intervention.
A related line of work is interested in modeling effort and progress in terms of detecting wheel-spinning, defined as prolonged amounts of time students spend on a problem without achieving mastery \cite{zhang2019wheel,wan2015wheel}. Wheel-spinning work examines on progress by identifying when effort is unproductive because it fails to translate into learning gains. However, such work still focuses on opportunity-to-mastery sequences rather than calendar-time horizons. 
On the other hand, predictions tasks are also common in MOOC and VLE settings, where models forecast dropout or stopout using aggregated clickstream features \cite{dekker2009dropout,wan2017moocdropout,gardner2019moocdropout}. However, these tasks are also generally classification problems centered on retention rather than continuous behavioral effort or learning progress within an ITS. Classic learning-rate and difficulty models (e.g., AFM, BKT, and related extensions) estimate latent learner and item parameters from practice histories, but they are typically not framed for supervised forecasting of observable engagement indicators \cite{cen2006,corbett1994knowledge}.}

\revisiontwo{At the same time, prior work has shown that effort-related measures—particularly time-on-task—can meaningfully predict learning outcomes, even when imperfect. For example, prior EDM studies have used time-on-task as measure of engagement to predict achievement and learning gains \cite{ritter2013timeontask}. Although time-on-task has known limitations and should not be treated as a perfect proxy for learning \cite{jla_time_on_task}, it remains practically valuable because it is simple and directly observable by tutors and students. For example, middle school students find it easier to express goals in time-on-task minutes than in progress metrics such as skill mastery \cite{peng2024homework}. Many educational platforms and vendors also recommend minimum weekly usage targets (e.g., 30 minutes) \cite{grimaldi2022estimating}, and in settings without detailed skill models, time-based measures may be the only consistently available indicator of effort (and arguably more comparable than the number of completed problems, which may vary in complexity across platforms). Therefore, defining effort as minutes practiced provides a straightforward, usable signal for week-to-week decision support.}

\revisiontwo{While prior work on affect detection, wheel-spinning, and weekly dropout prediction offers methodological inspiration (e.g., supervised machine learning (ML) models) \cite{hany2021performancepred}, these approaches do not address the same prediction target, temporal granularity, or prediction environment (ITS). In contrast, we introduce engagement forecasting as a supervised, week-ahead regression task within an ITS predicting two observable outcomes: minutes practiced (effort) and new skills mastered (progress). 
\revisionrd{This framing allows engagement forecasting to consider broader student context across time rather than step-level details, which practically situates predictions as week-to-week decision-support signals for tutors and students.
Furthermore, through interpretability analysis, continuous prediction of weekly student trajectory provides theoretical insights into longer-term predictive factors (e.g. student behaviors over several weeks) that prior EDM works with fine-grained measures rarely look at.}
}

\revisiontwo{Additionally, as prior prediction works are not directly applicable in our forecasting setup, we borrow work from Adams et al.~\cite{2x2} in the adjacent health behavior domain to provide a reasonable, non-naive baseline against our ML models. Adams et al.’s algorithm recommends a daily step target by looking at a person’s last 9 days of step counts and setting today’s target to the 60th percentile. However, while this heuristic algorithm is easily implementable, its transferability across domain (health to education) and time granularity (days to weeks) is an open question, and its heuristic nature may provide insight into whether or not statistical features actually bring more predictive power.}

\section{Dataset}

We focus on ITS as a seminal paradigm of adaptive learning systems \cite{vanlehn2011relative}. We use a longitudinal dataset available upon request \cite{largedataset} from students using web-based ITS for middle school mathematics. The dataset comprises activity logs collected from 425 students over 39 weeks (October 2010 to May 2012), capturing 2.9 million student-problem interactions. Students engaged with the system for an average of 39.8 $\pm$ 14.7 minutes per week, solving an average of 24.7 $\pm$ 13.1 problems weekly. Individual student-week observations (n = 11,710) showed considerable variability in engagement, with weekly practice time ranging from zero to over 6 hours (mean = 43.2 ± 25.9 minutes per student-week). \revision{The raw dataset contains 33 columns including: anonymized student identifiers, precise timestamps, interaction durations, learning outcomes, knowledge components, opportunity counts, and problem identifiers \cite{largedataset}. This event-level granularity enables aggregation of individual interactions into weekly summaries}. \revision{While our dataset is from 2012, it remains relevant to modern ITS usage context. In particular, fundamental tutor features in ITS remain the same (adaptive step-level guidance with feedback and hints, adaptive problem selection \cite{vanlehn2006behavior}). Data format also remains the same (e.g., through DataShop log data standards \cite{stamper2010datashop}).}

The unit of analysis for our engagement forecasting task is individual student-weeks. ITS automatically tracks two key outcomes: (1) time spent on practice problems and (2) whether related skills were mastered. We generate weekly ground-truth data for both minute and skill targets by: (1) aggregating time and skill data by ISO weeks (Monday--Sunday), creating identifiers such as ``2011-W23;'' (2) summing \revision{interaction durations} per student-week to create the first prediction target \textit{minutes practiced per week}; (3) Tagging a skill (knowledge component) as mastered when the AFM predicts proficiency $>$0.95 for a student-skill pair \cite{cen2006}; (4) Counting the number of new skills mastered per student-week to create the second prediction target \textit{new skills mastered per week}; (5) Filtering missing and invalid values and removing outliers with Tukey's Fence ($k=1.5$).

We describe the four categories of features we created below. The detail of the entire feature set is described in Table \ref{tab:features}.

\subsection{AFM-derived Features}

We adopt the AFM to derive interpretable \revisiontwo{features with respect to the learners} \cite{cen2006}. AFM models the probability that student $i$ answers a problem step $j$ correctly as:
\[
\Pr(Y_{ij}=1) = \mathrm{logit}^{-1} \left( \theta_i + \sum_{k=1}^{K} q_{jk}\,\beta_k + \sum_{k=1}^{K} q_{jk}\,\gamma_k T_{ik} \right),
\]
where $Y_{ij}$ is student $i$’s response on step $j$, $\theta_i$ represents student ability, $\beta_k$ is skill easiness, $\gamma_k$ is the learning rate for skill $k$, $T_{ik}$ is the number of prior opportunities student $i$ has had on skill $k$, and $q_{jk} \in \{0,1\}$ indicates whether step $j$ requires skill $k$ \cite{cen2008}. AFM explains performance as three additive components: (1) baseline student ability ($\theta_i$), (2) learning gained with additional opportunities ($\gamma_k T_{ik}$), and (3) the easiness of practiced skills ($\beta_k$). Because these terms are interpretable, we map them directly into features for engagement forecasting. \revision{Specifically, we derive \textit{student ability} from $\theta_i$,
\textit{student learning rate} from $\gamma_k T_{ik}$, and
\revisiontwo{\textit{student week difficulty}} from the negative weighted average of $\beta_k$ across skills practiced in one student-week, weighted by the number of encounters $T_{ik}$.} We recompute AFM parameters on a rolling five-week window, extracting updated student-level ability, learning rate, and difficulty values each week.

\subsection{Other Feature Groups}

Beyond AFM, we engineered additional feature groups to capture complementary aspects of
student engagement and performance. In general, we categorize the features into four groups for reference in analysis: \revisiontwo{(1) AFM-style learner features, (2) engagement activity, (3) \revisionrd{practice }gaps, and (4) prior achievement.}

\textit{Engagement Activity.}
We include basic engagement measures (minutes practiced, problems solved, opportunities, and
skills along with their lagged values. From these sequences, we derive changes, averages,
and variability statistics. These features serve as intuitive baselines, capturing both immediate
performance and historical trends.

\textit{\revisionrd{Practice }Gaps.}
Zeros in weekly minutes are treated as informative rather than missing values. Because 98\% of
students exhibit at least one temporal gap, and the average student has four gaps, we encode gap
patterns through features such as recent gaps, time since the last gap, and cumulative gap count.
These indicators capture disengagement episodes that could influence future engagement.

\textit{Prior Achievement.}
Prior achievement is a strong predictor of future learning outcomes
\cite{PriorAchPaper1}. We therefore include features capturing initial
performance (quartile of early skills), stability of performance (consistency score), and
improvement from early to later weeks. These serve as proxies for student preparedness and
trajectory.

\begin{table}[t] 
\centering
\caption{Overview of feature groups and individual features.}
\label{tab:features}
\small 
\begin{tabularx}{\columnwidth}{l X}
\toprule
\textbf{Feature} & \textbf{Description} \\
\midrule
\multicolumn{2}{l}{\textbf{AFM}} \\
student\_ability & Baseline performance \\
student\_learning\_rate & Improvement per opportunity \\
student\_week\_difficulty & Weighted average difficulty of done skills \\
\midrule
\multicolumn{2}{l}{\textbf{Engagement Activity}} \\
minutes\_current & Minutes practiced this week \\
minutes\_lag{k} & Minutes practiced up to $k$ weeks back \\
problems\_current & Problems solved this week \\
problems\_lag{k} & Problems solved up to $k$ weeks back \\
opportunities\_curr. & Opportunities tried this week \\
opportunities\_lag{k} & Opportunities tried up to $k$ weeks back \\
skills\_current & Skills estimate this week \\
skills\_lag{k} & Skills estimates up to $k$ weeks back \\
recent\_change\_* & Week-to-week and average changes (minutes, problems, skills) \\
minutes\_stats & Summary statistics (mean, std, range, iqr) of minutes practiced \\
problems\_stats & Summary statistics (mean, sum, std) of problems solved \\
\midrule
\multicolumn{2}{l}{\textbf{\revisionrd{Practice }Gaps}} \\
has\_recent\_gap & Indicator of zero-minute weeks recently \\
weeks\_since\_gap & Number of weeks since the last zero \\
gap\_count & Total number of zero-minute weeks \\
\midrule
\multicolumn{2}{l}{\textbf{Prior Achievement}} \\
start\_quartile & Quartile of initial skills \\
consistency\_score & Stability of skills relative to its mean \\
improvement & Change in skills from early to later weeks \\
\bottomrule
\end{tabularx}
\end{table}

\section{Methods}
Our methods comprise \revision{four} steps. First, we describe our ML models and their training procedure. Second, we define model evaluation metrics and the model selection process for engagement forecasting that can most generalize out of sample. Third, we describe \revision{interpretability }analysis conducted \revision{on }the four feature groups. \revision{Finally, we describe the protocol for carrying out our exploratory case study.}

\subsection{Model Training}
\label{sec:model_training}

\revision{For each student $i$ and week $t$, we construct a feature vector $x_{i,t}$ (described in Table~\ref{tab:features}) from the student's interaction history up to week $t$. We train separate models to predict next-week outcomes:
\begin{gather*}
y^{(\text{min})}_{i,t+1} = \text{minutes practiced in week } t+1 \\
y^{(\text{skill})}_{i,t+1} = \text{new skills mastered in week } t+1.
\end{gather*}
Thus, each sample is a one-step forecast, and sequential forecasts across weeks yield predicted engagement trajectories.}

\revision{We benchmark fifteen predictors spanning heuristic and supervised approaches. As simple temporal baselines, we implement: (1) last week’s value carried forward; (2) median across all past values for the same student; (3) median across non-zero past values; (4) mean across all past values; and (5) mean across non-zero past values.} The “non-zero” variants follow prior design by Adams et al.~\cite{2x2} and potentially avoid ambiguity: a zero entry may indicate either disengagement or simple absence. These baselines establish a lower bound for performance with naive statistical heuristics. We also adapt \revision{three percentile-based heuristics from prior work }in health behavior change \cite{2x2,dashboardgoalsetting}. These set the target at the (6) 50th, (7) 60th, or (8) 70th percentile of the most recent nine non-missing values, enabling us to examine whether the original 60th-percentile assumption holds when transferred from health to education. \revision{Because the original algorithm requires sufficient history}, we design a staged initialization procedure. In week 1, the \revision{prediction} is set to the dataset average. In week 2, the \revision{prediction} equals the student’s week-1 performance plus the gap between the dataset’s 50th and 60th percentile in week 1. From week 3 onward, until sufficient history is available, the \revision{prediction} is updated as 60\% of the way toward the highest of the student’s previous values. This approximates “somewhat above average” performance under limited history, aligning with \revision{engagement theory} \cite{ZPD}.

Beyond these eight baselines, we benchmark seven supervised learning methods that leverage \revision{the full engineered feature set}: (9) Random Forest (RF), (10) XGBoost (XGB), (11) Ridge regression, (12) LASSO regression, (13) linear regression (LR), (14) multilayer perceptron (MLP), and (15) long short-term memory (LSTM) network. This suite of models was chosen to span a spectrum of inductive biases---linear vs. nonlinear, tree-based vs. neural, memoryless vs. sequential---thereby probing which paradigms best forecast student engagement. All implementations use standard off-the-shelf libraries to ensure reproducibility, and the full pipeline is available in \revisioncamera{a public} repository\footnote{\revisioncamera{https://github.com/EricQiu6/EngagementForecasting}}.

\subsection{Model Evaluation}

\revision{We evaluate generalization to new students by using student-level splits. We allocate 70\% of students for model development and hold out the remaining 30\% of students for final testing, ensuring no student appears in both sets. Within the 70\% development set, we conduct 5-fold time-series split for model selection. For each student, predictions are produced in chronological order, always using information available up to the current week.}

\revision{We report MAE for interpretability. We also report relative error reduction:
$$
\Delta_{\%} = \frac{\mathrm{MAE}_{\text{compared model}}-\mathrm{MAE}_{\text{model}}}{\mathrm{MAE}_{\text{compared model}}}\times 100\%.
$$
To test significance, we estimate 95\% confidence intervals via paired bootstrap resampling over students ($B=10{,}000$).
}

To answer \textbf{RQ1}, we evaluate models across four settings for each model hyperparameter and five history window lengths \revisiontwo{(the number of past weeks a model can see)} to (1) investigate how well ML approaches forecast engagement, and (2) provide empirical analysis of what configurations and model families affect the performance gains of ML models. We then select the best-performing configurations for subsequent analysis. To answer \textbf{RQ2}, we compare weekly performance trends of our best ML model against the three Adams predictors \revision{on the remaining 30\% holdout set}. We examine whether the 60th percentile assumption aligns well with student behavior trends and analyze where data-driven models excel in cases where Adams do not generalize.

\subsection{Feature Importance Analysis}

To address \textbf{RQ3}, \revisiontwo{we use feature importance measures to examine which information derived from ITS logs is most indicative of future engagement levels.} For linear models (Ridge, LASSO, Linear Regression), importance is the absolute value of coefficients, indicating how strongly each feature relates to the target \cite{hoerl1970ridge, tibshirani1996regression, seber2003linear}. For tree ensembles (Random Forest, XGBoost), importance comes from impurity-based criteria, i.e., how much each feature reduces prediction error in splits \cite{ho1995random, chen2016xgboost}. To make results comparable, we average importances across validation folds and normalize them to sum to one. We then group features into the four categories and compute average weights for each group. This shows whether predictive gains come mainly from AFM-style learner parameters, engagement activity, practice gaps, or prior achievement.

On top of attribution, we conduct ablation study on the models to test whether feature groups are \emph{necessary}. For each group, we remove it and retrain the models, holding all other conditions fixed. We then compare MAE to the full model, estimating contribution of the excluded group. Bootstrap resampling across students provides confidence intervals and significance tests for these error differences.

\subsection{\revision{Studying Tutors' Use Of Signals}}
\begin{figure*}[t]
    \centering
    \begin{subfigure}[t]{0.49\linewidth}
        \centering
        \includegraphics[width=\linewidth]{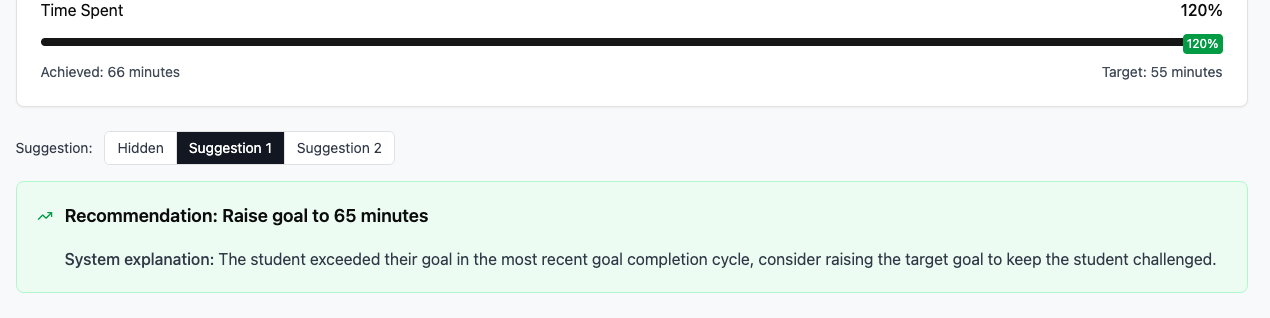}
        \Description{A dashboard showing a student's time spent on a task. A horizontal progress bar is filled beyond the target, labeled ``120\%'' Below it, text reads ``Achieved: 66 minutes'' and ``Target: 55 minutes.'' A suggestions section is visible with ``Suggestion 1'' selected. A green-highlighted recommendation box states: ``Recommendation: Raise goal to 65 minutes.'' Below, a system explanation reads: ``The student exceeded their goal in the most recent goal completion cycle, consider raising the target goal to keep the student challenged.''}
        \caption{}
        \label{fig:dashboard1a}
    \end{subfigure}\hfill
    \begin{subfigure}[t]{0.49\linewidth}
        \centering
        \includegraphics[width=\linewidth]{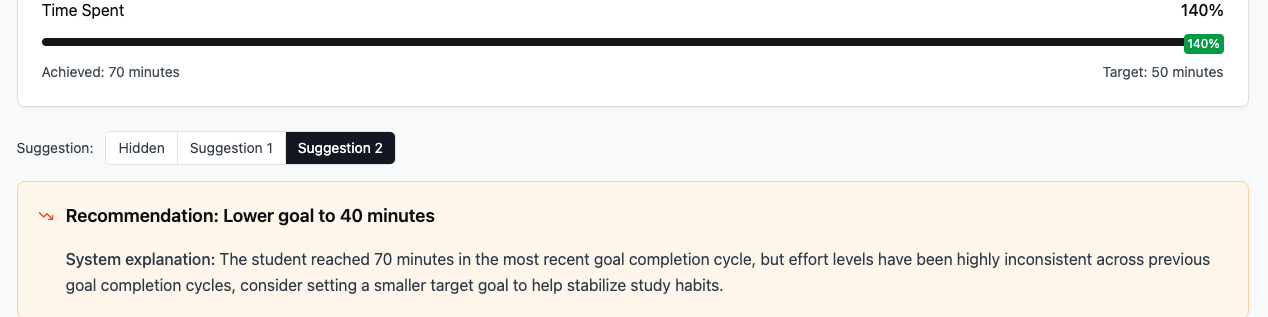}
        \Description{A dashboard showing a student's time spent on a task. A horizontal progress bar is filled well beyond the target, labeled ``140\%.'' Below it, text reads ``Achieved: 70 minutes'' and ``Target: 50 minutes.'' A suggestions section is visible with ``Suggestion 2'' selected. An orange-highlighted recommendation box states: ``Recommendation: Lower goal to 40 minutes.'' Below, a system explanation reads: ``The student reached 70 minutes in the most recent goal completion cycle, but effort levels have been highly inconsistent across previous goal completion cycles, consider setting a smaller target goal to help stabilize study habits.''}
        \caption{}
        \label{fig:dashboard1b}
    \end{subfigure}
    \caption{\revisiontwo{(a) \textbf{Intuitive goal-setting scenario (minutes practiced).} Example of an intuitive case that raises goal because student exceeded prior week's goal. (b) \textbf{Counter-intuitive goal-setting scenario (minutes practiced).} Example of a counter-intuitive case that lowers goal even though the student exceeded prior week's goal, citing the consistency feature as reason.}}
    \label{fig:case_study_side_by_side}
\end{figure*}

\revision{
We conducted a \revisionrd{semi-structured qualitative user interview case} study to outline how \revisiontwo{the model signals we investigated (e.g. learning rate, difficulty, consistency)} may inform real-world tutoring. The study protocol was approved by our Institutional Review Board (IRB), and all participants provided informed consent and received compensation for their time. Participants (N = 8 college students doing part-time K–12 math tutoring online) interacted with a goal-setting dashboard across four scenarios. \revisioncamera{This sample size aligns with comparable educational design research \cite{XAI} and was guided by qualitative saturation, as few new themes emerged in later interviews.} Each scenario varied by goal type (minutes practiced vs. skills mastered) and alignment with tutor intuition (intuitive vs. counter-intuitive) to examine how tutors interpret and respond to explainable features that may conflict with the tutors' initial judgement. \revisiontwo{The dashboard displayed recent goal cycles, achieved vs. target progress, and a system recommendation panel with three settings: (1) no recommendation, (2) recommendation of a goal value only, and (3) recommendation with a justifying explanation that cites relevant features extracted from our statistical models. We designed this three-variants setup to respectively (1) capture each tutor’s baseline judgment without any recommendation influence, (2) measure how a recommended value alone shifts tutor decisions, and (3) isolate the incremental effect of the explainable features beyond the system recommendation itself. Scenarios and explanation variants were presented in randomized order. See Fig.~\ref{fig:case_study_side_by_side} for example dashboard interfaces.} The study was descriptive and did not measure learning outcomes; it aimed to characterize tutor interpretation and decision behavior when explanations reference the interpretable features, thus clarifying whether the forecasting signals could be helpful and used in downstream tutoring scenarios. We collected qualitative think-aloud data to capture tutors’ mental models and decision-making processes.
}

\begin{table}[t]
\centering
\caption{MAE summary for minutes and skills. Lower is better. Top feature-augmented model (XGBoost) vs.\ average of eight baselines.}
\label{tab:mae_summary}
\setlength{\tabcolsep}{4pt}
\renewcommand{\arraystretch}{1.05}
\footnotesize
\begin{tabularx}{\columnwidth}{@{}Xcc@{}}
\toprule
\textbf{Target} & \textbf{Baseline MAE} & \textbf{Top Model MAE ($\Delta_{\%}$)} \\
\midrule
Minutes/week    & \emph{10.37} & \emph{8.05 ($-22.4\%$)} \\
New skills/week & \emph{5.21}  & \emph{3.51 ($-32.6\%$)} \\
\bottomrule
\end{tabularx}
\end{table}

\section{Results}
\label{sec:results}

We first test how accurately feature-based models forecast next-week engagement and which settings affect these gains. Next, we check whether percentile heuristics from other domains transfer to weekly K–12 engagement by comparing Adams baselines against our best ML model. \revision{Then}, we assess which feature categories drive performance by analyzing importance scores and ablation results for AFM-based features, engagement activity, \revisionrd{practice} gaps, and prior achievement. \revision{Finally, we examine the eight case study sessions to interpret our explainable features in context of downstream tutor understanding during goal setting.}

\subsection{RQ1: forecasting effort and progress}
\label{sec:rq1}

\textit{Performance gains over baselines.}
Data-driven ML models substantially outperform heuristic baselines (last value, mean/median variants, and Adams-style methods \cite{2x2,dashboardgoalsetting}). On average, error falls by \textbf{22\%} for minutes and \textbf{33\%} for skills. Bootstrap confidence intervals on paired percentage differences exclude zero across linear, tree, and neural families, establishing that improvements are statistically significant. To better understand where these gains come from, we next compare across \revisiontwo{model family} and hyperparameters.

\revisiontwo{\textit{Family-level differences.}}
As shown in Table \ref{tab:RQ1_family_tests}, \revisiontwo{family-level} comparisons reveal only modest variation. \revision{For the \textbf{minutes} target, linear vs.\ tree shows a non-significant MAE gap (mean diff $=0.16$; 95\% CI $[-0.07,\,0.40]$; $p=.318$; Cliff's $\delta=0.08$). However, while linear and tree models outperform neural nets at shorter histories, this gap narrows to non-significance as history length increases past 20 weeks, consistent with neural nets’ higher data requirements. For the \textbf{skills} target, all pairwise comparisons yield negligible differences (e.g., linear--tree mean diff $=0.02$, CI $[-0.13,\,0.16]$, $p=.362$, $\Delta_{\%}=0.03$), with effect sizes $|\delta|\leq 0.07$.} Thus, \revisiontwo{model family} matters somewhat for minutes under short histories, but practical differences vanish for skills.

\begin{table}[t]
\centering
\caption{Pairwise family comparisons by target (MAE and $\Delta_{\%}$ between the row model and the comparator). Lower is better. \textbf{Bold} marks significance. $^{*}$ $p < .05$; $^{**}$ $p < .01$}
\label{tab:RQ1_family_tests}
\small
\setlength{\tabcolsep}{2.5pt} 

\begin{subtable}{\columnwidth}
\centering
\caption{Skills}
\begin{tabularx}{\columnwidth}{l X c c c c}
\toprule
Family & Models & MAE & vs.Tree & vs.Linear & vs.Neural \\
\midrule
Linear & LR,Ridge,LASSO & 3.60 & 0.02 & -- & -0.11 \\
Tree   & RF, XGB          & 3.50 & -- & -0.02 & -0.13 \\
Neural & MLP, LSTM        & 3.68 & 0.13 & 0.11 & -- \\
\bottomrule
\end{tabularx}
\end{subtable}

\vspace{1em}

\begin{subtable}{\columnwidth}
\centering
\caption{Minutes}
\begin{tabularx}{\columnwidth}{l X c c c c}
\toprule
Family & Models & MAE & vs.Tree & vs.Linear & vs.Neural \\
\midrule
Linear & LR,LASSO,Ridge & 9.04 & 0.16 & -- & \textbf{-1.72**} \\
Tree   & RF, XGB          & 8.76 & -- & -0.16 & \textbf{-1.88**} \\
Neural & MLP, LSTM        & 9.46 & \textbf{1.88**} & \textbf{1.72**} & -- \\
\bottomrule
\end{tabularx}
\end{subtable}
\raggedright
\end{table}

\textit{Hyperparameter sensitivity.}
Within each family, performance varies little across hyperparameter settings. Friedman tests blocking by window$\times$fold show low and nonsignificant concordance (Kendall’s $W<.05$), with no configuration emerging as a consistent winner. Accuracy is robust to parameter choice; most reasonable configurations yield errors within a narrow range.

\textit{Takeaway.}
Data-driven models greatly outperform baselines by 22–33\% (Table \ref{tab:mae_summary}). However, the decisive factor for performance gains is not which family or hyperparameter is chosen. Linear, tree, and neural families all reach similar accuracy given sufficient history, and hyperparameters rarely alter the outcome. On practical grounds, we adopt \textbf{XGBoost} as a default model: it performs best empirically, trains efficiently, and provides directly interpretable attributions for downstream analyses.

\subsection{RQ2: do heuristic rules generalize}
\label{sec:rq2}

We evaluate Adams-style percentile heuristics \cite{2x2,dashboardgoalsetting} against our chosen feature-based predictor XGBoost. These heuristics predict engagement through a fixed percentile (e.g., 50th, 60th, 70th) of past performance. The $>$50th-percentile rules assume \emph{continuous improvement}: \revision{if learners repeatedly meet their targets, the prediction will continue to increase.}

\begin{figure*}
    \centering
    \includegraphics[height=6cm]{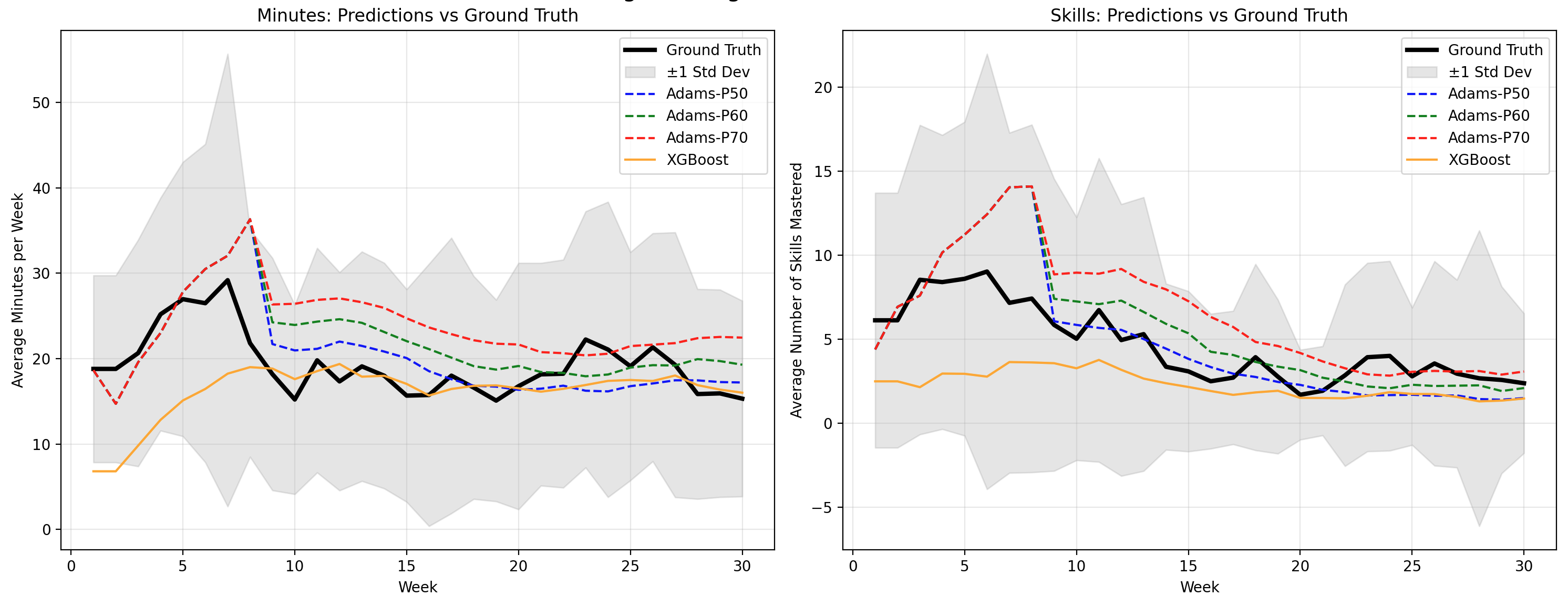}
    \Description{Two line charts compare weekly ground truth with model predictions for average minutes practiced and skills mastered. Grey bands around the lines show 1 standard deviation.}
    \caption{Average minutes and skills per week across all students plotted against the predictions of different models, also averaged across students (XGBoost, 50th percentile, 60th percentile, and 70th percentile variants of Adams predictor). Grey band is the standard deviation spread of the ground truth data.} 
    \label{fig:rq2_trend}
\end{figure*}

Fig.~\ref{fig:rq2_trend} shows that this assumption does not hold in our setting. Effort (minutes practiced per week) engagement stabilizes after an early surge, and progress (new skills mastered per week) rises initially but then declines mid-semester. As a result, Adams predictors systematically overshoot: P60 and P70 in particular perform 4--33\% worse than P50 across targets, and averaging across percentiles further degrades accuracy. Table~\ref{tab:rq2_summary} shows full performance results.

XGBoost reduces MAE by 20--30\% relative to Adams-P50, and by larger margins relative to Adams-P60 and P70 (Table~\ref{tab:rq2_summary}). Still, its performance in the earliest weeks is weaker, and for the skills target, it tends to underpredict systematically. The spread bands in Fig.~\ref{fig:rq2_trend} help explain this pattern: the dataset exhibits far greater instability early in the semester. This higher variance reduces the stability of any predictor and plausibly explains why XGBoost lags in early weeks and underestimates peaks in skills. Once trajectories stabilize starting week 9, XGBoost outperforms Adams predictors by even larger margins and tracks both effort and progress levels much more closely.

\textit{Takeaway.} Percentile-based heuristics from other domains do not transfer cleanly to weekly K--12 practice. Their built-in continuous improvement assumption becomes inaccurate once growth slows, leading to systematic overprediction. A feature-based model like XGBoost not only avoids this pitfall but adapts better to shifting regimes in the data, providing substantially more accurate predictions given sufficient data.

\begin{table*}[t]
\centering
\caption{Performance comparison of XGBoost vs. Adams variants  (P50, P60, P70) for minutes and skills prediction. 
MAE (lower is better). $\Delta_{\%}$ values are relative to Adams-P50 and to the average of Adams variants.}
\label{tab:rq2_summary}
\begin{subtable}{\textwidth}
\centering
\caption{Minutes prediction}
\resizebox{\textwidth}{!}{%
\begin{tabular}{lccccccccc}
\toprule
\multirow{2}{*}{\textbf{Model}} & 
\multicolumn{3}{c}{\textbf{Final Week}} & 
\multicolumn{3}{c}{\textbf{Entire Sequence}} & 
\multicolumn{3}{c}{\textbf{Starting Week 9}} \\
\cmidrule(lr){2-4}\cmidrule(lr){5-7}\cmidrule(lr){8-10}
 & MAE & $\Delta_{\%}$ vs P50 & $\Delta_{\%}$ vs Avg Adams 
 & MAE & $\Delta_{\%}$ vs P50 & $\Delta_{\%}$ vs Avg Adams 
 & MAE & $\Delta_{\%}$ vs P50 & $\Delta_{\%}$ vs Avg Adams \\
\midrule
Adams-P50  & 10.1905 & 0.0\% & -5.2\% & 10.0119 & 0.0\% & -4.6\% & 9.5142 & 0.0\% & -7.6\% \\
Adams-P60  & 10.6971 & +5.0\% & 0.0\% & 10.4442 & +4.3\% & 0.0\% & 10.1037 & +6.2\% & 0.0\% \\
Adams-P70  & 11.8806 & +16.6\% & +11.1\% & 11.0962 & +10.8\% & +6.2\% & 10.9928 & +15.5\% & +8.8\% \\
XGBoost     & \textbf{7.8525} & \textbf{-22.9\%} & \textbf{-26.6\%} & \textbf{8.7210} & \textbf{-12.9\%} & \textbf{-16.5\%} & \textbf{7.2946} & \textbf{-23.3\%} & \textbf{-27.8\%} \\
Avg Adams   & 10.9227 & +7.2\% & 0.0\% & 10.5174 & +5.0\% & 0.0\% & 10.2036 & +7.2\% & 0.0\% \\
\bottomrule
\end{tabular}}
\end{subtable}

\vspace{1em}

\begin{subtable}{\textwidth}
\centering
\caption{Skills prediction}
\resizebox{\textwidth}{!}{%
\begin{tabular}{lccccccccc}
\toprule
\multirow{2}{*}{\textbf{Model}} & 
\multicolumn{3}{c}{\textbf{Final Week}} & 
\multicolumn{3}{c}{\textbf{Entire Sequence}} & 
\multicolumn{3}{c}{\textbf{Starting Week 9}} \\
\cmidrule(lr){2-4}\cmidrule(lr){5-7}\cmidrule(lr){8-10}
 & MAE & $\Delta_{\%}$ vs P50 & $\Delta_{\%}$ vs Avg Adams 
 & MAE & $\Delta_{\%}$ vs P50 & $\Delta_{\%}$ vs Avg Adams 
 & MAE & $\Delta_{\%}$ vs P50 & $\Delta_{\%}$ vs Avg Adams \\
\midrule
Adams-P50  & 2.7255 & 0.0\% & -13.2\% & 4.4260 & 0.0\% & -10.3\% & 3.4020 & 0.0\% & -15.0\% \\
Adams-P60  & 3.1333 & +15.0\% & +0.5\% & 4.7458 & +7.2\% & +0.9\% & 3.8380 & +12.8\% & +0.3\% \\
Adams-P70  & 3.5843 & +31.6\% & +15.9\% & 5.2541 & +18.7\% & +11.5\% & 4.5312 & +33.2\% & +18.3\% \\
XGBoost     & \textbf{2.2059} & \textbf{-19.1\%} & \textbf{-31.5\%} & \textbf{3.5099} & \textbf{-20.7\%} & \textbf{-22.4\%} & \textbf{2.5698} & \textbf{-24.5\%} & \textbf{-32.6\%} \\
Avg Adams   & 3.1477 & +15.5\% & 0.0\% & 4.8086 & +8.6\% & 0.0\% & 3.9237 & +15.3\% & 0.0\% \\
\bottomrule
\end{tabular}}
\end{subtable}
\end{table*}

\subsection{RQ3: which features are most indicative}
\label{sec:rq3}

\begin{figure}[t]
\centering
\includegraphics[width=\columnwidth]{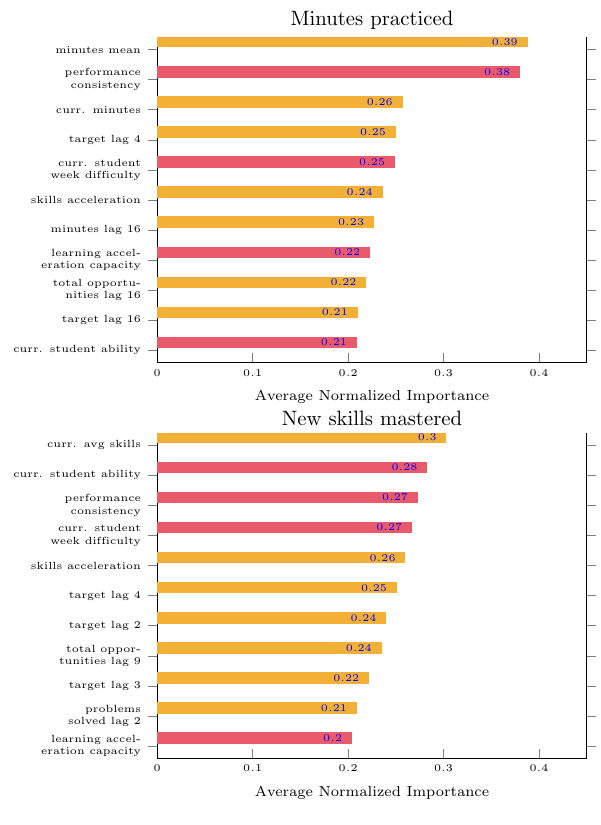}
\Description{
Two horizontal bar charts showing feature importance. The top chart, titled “Minutes practiced,” ranks features such as minutes mean, performance consistency, current minutes, and target lag 4. The bottom chart, titled “New skills mastered,” shows current average skills as most important, followed by current student ability, performance consistency and week difficulty. The x-axis in both charts is “Average Normalized Importance.”
}
\caption{Feature-importance rankings. Red: learner-centered; orange: recent activity. Top: minutes; Bottom: skills.}
\label{fig:rq3_topk}
\end{figure}

The top–$k$ feature plots (Fig.~\ref{fig:rq3_topk}) show two patterns, with calibrated strength. For \textbf{effort} (minutes practiced), the top features are primarily \revisiontwo{driven by \emph{recent engagement activity}:} \revision{\textit{minutes mean}, \textit{current minutes per week}, and a short lag \textit{target lag 4} take }three of the top four slots; additional lag features also appear in the top–11 (\textit{minutes per week lag 16}, \textit{target lag 16}, \textit{total opportunities lag 16}). However, stability across model families is low: linear and tree models rank features differently (Kendall’s $\tau \approx 0.28$), and category-level agreement is nearly zero ($\tau < 0.01$). In short, effort is driven mainly by \textit{\revisiontwo{recent engagement activity}}, with smaller contributions from learner modeling.

For \textbf{progress} (skills mastered), the key features are \revisiontwo{\textit{learner-centered}} indicators:
\textit{current student ability}, \textit{performance consistency score}, and \revisiontwo{\textit{{current student week difficulty}}} occupy three of the top four positions. Crucially, this ordering is more stable across model families: per–feature consistency is moderate ($\tau \approx 0.37$) and the category–level consistency is strong ($\tau \approx 0.67$). We therefore conclude that progress engagement is shaped mainly by {\textit{learner-centered signals}} that are robust to \revisiontwo{choice of model family}, whereas effort reflects a \revisiontwo{recent-activity}-driven but \revisiontwo{family-sensitive} mixture of \revisiontwo{engagement activity} and learner modeling.

Ablation tests (Table \ref{tab:ablation_deltapct}) confirm these patterns. For \textbf{minutes}, removing AFM, \revisionrd{practice} gaps, or prior achievement features causes only small losses ($<1\%$), while recent \revisiontwo{engagement activity} captures nearly all accuracy (differences versus the full set are $< 0.3\%$). For \textbf{skills}, every group contributes: dropping \revisiontwo{engagement activity} features yields the largest loss ($>1\%$), but AFM, gaps, and prior achievement features also provide meaningful differences (0.6--0.8\% each). Taken together, these results align with conclusions drawn from the feature rankings: effort forecasts are recent-activity-led, while skills forecasts rely more on \revisiontwo{learner-centered} signals.

\begin{table}[t]
\centering
\caption{Ablation results. MAE (lower is better) and relative change $\Delta_{\%}$ vs.\ \emph{Full}. 
Bold $\Delta_{\%}$ indicates bootstrap CI on the paired difference excludes $0$ (significant).}
\label{tab:ablation_deltapct}
\setlength{\tabcolsep}{3.5pt} 
\renewcommand{\arraystretch}{1.05}
\footnotesize
\begin{tabularx}{\columnwidth}{@{}Xcccc@{}}
\toprule
\multirow{2}{*}{\textbf{Condition}} & \multicolumn{2}{c}{\textbf{Skills}} & \multicolumn{2}{c}{\textbf{Minutes}} \\
\cmidrule(lr){2-3}\cmidrule(lr){4-5}
 & MAE & $\Delta_{\%}$ & MAE & $\Delta_{\%}$ \\
\midrule
\textit{Full (reference)}        & 4.1904 & 0.00\%          & 9.3067 & 0.00\% \\
Base features only          & 4.2151 & \textbf{0.59\%}  & 9.3226 & \textbf{0.17\%} \\
Base + activity             & 4.2355 & \textbf{1.08\%}  & 9.3082 & 0.02\% \\
Base + gaps                 & 4.1996 & \textbf{0.22\%}  & 9.3125 & 0.06\% \\
Base + prior                & 4.2173 & \textbf{0.64\%}  & 9.3188 & 0.13\% \\
Base + AFM                  & 4.2129 & \textbf{0.54\%}  & 9.3286 & \textbf{0.24\%} \\
All except activity         & 4.2300 & \textbf{0.95\%}  & 9.3088 & 0.02\% \\
All except gaps             & 4.2141 & \textbf{0.57\%}  & 9.3155 & \textbf{0.09\%} \\
All except prior            & 4.2239 & \textbf{0.80\%}  & 9.3111 & 0.05\% \\
All except AFM              & 4.2180 & \textbf{0.66\%}  & 9.3122 & \textbf{0.06\%} \\
\bottomrule
\end{tabularx}
\end{table}

\subsection{\revision{Case Study Results}}

\revision{We examined how eight tutors interacted with system generated recommendations and accompanying explanations that referenced distinct feature types identified in RQ3. Two independent coders conducted thematic analysis of the think-aloud data. We describe our derived patterns below.}

\revision{First, tutors consistently preferred rationales backed by multi-week historical data. One participant noted, “Past data… tendency… really convinces me." This aligns with our statistically driven prediction approach, suggesting that, beyond the prediction values, the derived explainable features themselves may be valuable in downstream settings to assist in tutoring decision-making.
}

\revision{Second, tutors' mental models diverged by target type in a way that mirrors our prediction target split. For effort (minutes), tutors emphasized routine formation and sustainable baselines, consistent with recency-driven engagement dynamics (e.g., lags, recent change, and variability). For progress (skills), tutors more often reasoned about learner and content properties (e.g., ability, consistency, and difficulty), which matches our feature-importance pattern where AFM-derived ability and week difficulty are central. 
}

\revision{Third, tutors expressed skepticism toward difficulty and ability claims unless the definition was clear. In addition, a few tutors raised concerns about sensitivity around student-facing explanations. For example, one participant remarked, ``If the words say… `low learning rate', I don’t think that’s a good thing to show kids.'' These critiques reinforce a key design implication for deploying model-derived signals as explainable evidence: even when predictions may be accurate, their impact depends on careful communication that preserves learner agency and avoids discouragement. 
}

\revision{Lastly, when qualifying tutor acceptance and trust, we find that participants generally prefer to take final control of the specific goal value. \revisiontwo{However, across the 2x2 scenario split by goal type (minutes practiced vs. skills mastered) and intuition (intuitive vs. counter-intuitive)}, all tutors are most influenced by the third explanation variant that provided additional feature-backed evidence. In particular, one response characterized the tutors' trust: ``if the system… takes the history of students, and it suggests based on that history, then I think I would be totally okay to use the suggestion of AI.'' 
\revisionrd{Together, these findings show that while tutors preferred to retain final say, they trusted the explainable features, finding the system helpful in designing plans with their student.}
These feedback suggest that explainable signals may have the potential to aid decision-making in human-AI tutoring.}

\revision{Overall, the tutor sessions help interpret our modeling results: the signals that are most predictive (recent history for effort; learner ability and difficulty for progress) are also the signals tutors find most helpful. \revisionrd{This alignment between the signal's predictive power and tutor reasoning suggests that engagement forecasting offers a foundation for utilizing explainable contextual information to support timely instructional decisions, which may in turn lead to more calibrated and consistent engagement plans for students.
} 
}

\section{\revision{Discussion}}
\revisionrd{The present study is motivated by a persistent challenge in ITS: students often do not practice consistently enough to fully realize the system’s potential learning gains, even within scheduled classroom time}
\cite{gurung2025starting, eames2026khan}. While EDM has made major progress in modeling knowledge and performance, comparatively less work has treated effort across weeks as primary analytic targets \cite{dekker2009dropout, mu2020wheel}. 
Our contribution is formalizing and benchmarking \textit{engagement forecasting} as a novel supervised learning task over ITS log data. \revisionrd{This empirical contribution (forecasting and feature analysis) makes patterns of sustained engagement visible and explainable, supporting its potential as a decision-support tool for teachers and students to construct more calibrated and personalized plans for effective and consistent engagement.}

The remainder of this section is organized by research question. For each RQ, we recap key results, then interpret its significance in the context of prior work, draw implications for EDM and educational practice, and outline directions for future work. \revisionrd{We conclude with analysis of our case study to connect engagement forecasting to real-world settings.}

\subsection{Forecasting suggests practical viability}

Answering \textbf{RQ1}, feature-driven models achieve sizeable reductions in forecasting error---approximately 22\% for minutes practiced and 33\% for new skills mastered---relative to multiple heuristic baselines \cite{2x2}. Notably, performance is similar across linear and tree-based families, and hyperparameter differences are small once sufficient history is available. This convergence suggests that, for engagement forecasting, data coverage and feature design may matter more than selecting a particular model class. 
Practically, \revisiontwo{this model insensitivity} is encouraging: systems can deploy relatively standard models (e.g., gradient-boosted trees or regularized regressions) and still obtain large improvements over heuristics, lowering barriers to adoption. Notably, although we demonstrate that our data-driven approach is successful at improving future effort and \revision{progress} prediction, we do not claim that this exact model or feature stack generalizes to other contexts due to varying engagement dynamics. Instead, we offer open-source code to apply our models to local data and ensure that features are available and meaningful with respect to the class dynamics \cite{gurung2025starting}.

\revisiontwo{Crucially, improved forecasting accuracy alone does not imply improved learning outcomes. Rather, week-ahead predictions are best interpreted as \textbf{decision-support signals}. For example, forecasts can flag students whose engagement may deviate from recent patterns, while still leaving room for instructors and tutors to situate the forecasts using contexts the system cannot access, such as classroom constraints, instructional goals, and students’ schedules. In this case, a tutor might discuss with a student any upcoming events that would get in the way of normal engagement. Such interaction helps refine engagement expectations and makes predictions more useful for week-to-week support.
}

\subsection{Transferred heuristics do not match}
Answering \textbf{RQ2}, percentile heuristics adapted from adjacent \revisiontwo{health behavior domains} \cite{2x2,dashboardgoalsetting} do not generalize cleanly to K--12 weekly trajectories. \revision{Percentile} rules that target above-median performance systematically overshoot, with P60 and P70 performing worse than P50 across targets, and our feature-driven model outperforming P50 by 20-30\%. 

Our analysis shows that such percentile heuristics does not reflect typical ITS usage patterns over a school year. Factors such as limited instructional time, schedule disruptions, and shifting curricular focus mean that consistently exceeding past performance is neither realistic nor always desirable \cite{gurung2025starting}. 
\revisiontwo{This observation seems consistent with prior work \cite{gurung2025starting, eames2026khan} showing large variations in how students use available time---the limited ability for a single heuristic to model all students may in part explain its lower prediction accuracy.}
In contrast, feature-based forecasting is more flexible: it can represent plateauing, regression, and regime shifts as a function of learner model (AFM), recent activity, inactivity patterns, and prior achievement. This flexibility is likely necessary for realistic decision support in classrooms with limited time and schedule disruptions.

\subsection{Effort and progress have distinct drivers}

Answering \textbf{RQ3}, our interpretability analyses reveal a consistent distinction between the two targets. \textbf{Effort} (minutes practiced) is driven by \revisiontwo{recent engagement activity}, while \textbf{progress} (new skills mastered) depends more strongly on \revisiontwo{learner-centered signals such as AFM-derived ability and difficulty, as well as consistency.} 
AFM features may not always be available in adaptive learning systems \cite{ITS2, aleven2025integratedplatformstudyinglearning}. Still, simplified measures of content difficulty or student ability estimated from past students' pass rates could still be used in other models. Alternatively, our ablation study suggests these features could also be removed while retaining viable model performance. 
In addition, data-driven modeling is weaker in the starting weeks for both targets. Nevertheless, since results show that XGBoost widens \revision{the performance gap against baselines }starting week 9, one mitigation strategy may be to employ the same initialization procedure in the starting weeks as we constructed the Adams predictors (see Section \ref{sec:model_training}) \cite{2x2, ZPD}.

\revisiontwo{The divergence between the sets of predictive signals for effort vs. progress seems theoretically counterintuitive, since learning progress ultimately arises through effort. While investigating a satisfactory explanation is beyond the scope of this work, we hypothesize that the definition of progress in terms of weekly skills mastered may be too coarse-grained. That is, the progress-effort relationship may be obscured by discrete skill mastery, whereas classical measurements of proficiency (e.g., AFM, BKT) operate at a step-level \cite{cen2006, corbett1994knowledge}. 
}\revisiontwo{Nevertheless, defining progress as skills mastered is practically valuable as it is more aligned with this work's intention to view predictions as week-to-week decision-support signals for tutors and students. Furthermore, the separation of targets is useful for practical downstream use cases. By comparing feature groups, we aim to isolate which signals most predict student persistence and to potentially translate these insights into improved ITS adaptivity---for example, adjusting difficulty in mastery-based problem selection receptive to student motivation and effort \cite{Xia2025OptimizingML}. These signals may also provide explanatory insights to learner and teacher feedback in downstream applications. We expand on this in the next case study section.}

\subsection{\revision{Implications for Explainable Interfaces}}

\revisiontwo{Across the case study, three observations were most consistent: (1) when judging recommendations, tutors placed high trust in system explanations backed by the explainable features when their definition and data source (weeks of personalized student data) were clear; (2) tutors reasoned differently about effort vs. progress in ways that mirror our target-specific feature drivers; and (3) tutors identified sensitivity concerns of learner-modeling claims (e.g., difficulty and ability) when recommendation becomes student-facing.}

\revision{Together, these findings suggest that \revisiontwo{the top features extracted in RQ3 (e.g. ability, difficulty, consistency)} are not only predictive inputs but also candidate explanatory evidence. 
We identify feature groups that map naturally onto the kinds of reasons tutors use, \revisiontwo{while also revealing that practical explainability may break down if the sources of the features are unclear (i.e. tutor confusion about definition of ability and difficulty). These findings suggest that grounded and clearly defined features may be practically compatible with existing tutor reasoning workflows, situating feature analysis as more than a post-hoc interpretability exercise.}}

\revision{While we find that the features themselves may provide helpful explanatory power to decision making in downstream contexts such as goal setting, the direct usage of the engagement prediction values in these applications should be further scrutinized.  For example, accurate week-ahead predictions might not determine appropriate goals. \revisiontwo{In particular, while a model may predict declining engagement accurately, it is not clear whether using declining value as goal recommendation is optimal for student motivation.} One promising direction of work could define and test explicit forecast-to-recommendation mappings that balance challenge and motivation while incorporating pedagogical constraints (e.g., schedule disruptions, shifting curricular focus, and teacher-configured limits). \revisiontwo{This would close the loop from prediction to intervention and enable in-situ evaluation of downstream outcomes, including tutor adoption, student acceptance, and learning gains.}}

\subsection{Limitations and Future Work}

\revisiontwo{A few limitations bound the generalizability of our findings. The study is restricted by}
a single ITS, subject domain, as well as platform-specific logging; some inputs underpinning our features (e.g., opportunity counters, fine-grained timestamps, skill tags for AFM) are not universal and may require proxies or simplified learner-state estimates when absent. External validation should therefore include cross-site replication and leave-site-out testing. \revisiontwo{In addition, we do not distinguish between classes that assign homework and those that do not, even though these contexts differ in how student agency and self-regulated learning operate \cite{SRL, SRL2}. Future work could explicitly examine how forecasting performance and interpretation differ across instructional contexts.}
\revision{Finally, an open design question is how to communicate forecasts effectively. Prior research on open learner models and explainable learning analytics suggests that transparent representations of learner state can support self-regulation \cite{bull2007smili,hooshyar2020open}. Translating forecasts into actionable feedback without discouraging learners requires careful design and  communication. \revision{While we motivate towards this direction with our case study,} such work is beyond the scope of this paper and may be a valuable next step.}

\revisioncamera{Several open questions about the prediction task itself also merit further investigation. Our contribution is the formulation and empirical validation of week-ahead engagement forecasting using standard ML methods. We leave algorithmic innovations on this prediction task for future works. In particular, further insights into which features are predictive and how accurate student engagement can be modeled may be valuable. Relatedly, both of our targets---minutes practiced and new skills mastered---are useful but imperfect proxies for effort and learning progress \cite{jla_time_on_task}. Alternative proxies---including problem attempts, session frequency, and active-vs.-idle time—were not directly compared in this work. As prior research show that students naturally articulate goals in both problem-count and time-duration terms \cite{peng2024homework}, a comparative study of proxy choice as the forecasting target is a natural next step.}

\revisiontwo{Looking forward, engagement forecasting may support a wider range of applications than goal setting alone. Potential uses include instructional pacing and estimating time-to-mastery under revised content—contexts where week-by-week forecasts could inform planning even without direct student-facing interventions. Evaluating such usages, and studying how forecasts interact with teacher-student decision-making, remains an important area for future research.}

\section{Conclusion}
\revisionrd{
Prior work in EDM has introduced statistical learning methods to predict important targets such as learning, affect, momentary disengagement, performance, and dropout. However, comparatively little work has focused on explainable forecasting of student effort and progress at a weekly timescale, aligned with practical instructional decision-making related to practice goal setting. In this paper, we introduce engagement forecasting as a novel prediction task in ITS and benchmark models for forecasting two observable outcomes: minutes practiced per week (effort) and new skills mastered per week (progress).
}

\revisionrd{
First, as a methodological contribution to EDM, we showed that statistical learning models are well suited to this task, yielding 20--30\% reductions in forecasting error relative to heuristic baselines transferred from other behavioral domains. Our results further demonstrate that data-driven models better capture non-monotonic patterns common in real classroom distributions.
Second, through interpretability analysis, we provided theoretical insight into the structure of engagement, showing that effort and progress are driven by qualitatively different features. Future effort is driven by recent engagement activity, whereas future progress depends more on learner-centered signals such as ability, difficulty, and consistency. These findings contribute insights into the features that learners need to consider to effectively calibrate their effort, a dimension rarely supported in adaptive learning systems.
Third, through a semi-structured tutor interview, we examined how these predictive signals are interpreted in downstream goal-setting contexts. We found that tutors reason differently about effort vs. progress in ways that align with our findings, suggesting that the signals identified through engagement forecasting are not only predictive but also meaningful for constructing more personalized plans for effective and consistent engagement.
While this study does not evaluate learning outcomes, it provides a foundation for future work linking engagement forecasts to classroom learning benefits, given prior evidence that time-on-task and related effort measures can be predictive of learning outcomes in tutoring software \cite{ritter2013timeontask}. This foundation enables future work that explicitly links forecasts to goal-setting strategies and instructional planning, evaluated on recommendation acceptance, persistence, and learning outcomes.
}

\section{\revisioncamera{Acknowledgments}}
\revisioncamera{This research was primarily funded by the Institute of Education Sciences (IES) of the U.S. Department of Education under award \#R305A220386 and was partially supported by the National Science Foundation under Award No. 2349558. This work was supported, in whole or in part, by the Bill \& Melinda Gates Foundation INV-068909. The conclusions and opinions expressed in this work are those of the author(s) alone and shall not be attributed to the Foundation. Under the Foundation’s grant conditions, a Creative Commons Attribution 4.0 Generic License has already been assigned to the Author Accepted Manuscript version that might arise from this submission.
We also thank Joyce Gill for her valuable editorial insights.}

%
\bibliographystyle{abbrv}
\bibliography{sigproc}  

\end{document}